\documentclass[conference]{IEEEtran}
\IEEEoverridecommandlockouts
\usepackage{cite}
\usepackage{amsmath,amssymb,amsfonts, amsthm}
\usepackage{algorithmic}
\usepackage{graphicx}
\usepackage{textcomp}
\usepackage{xcolor}
\newtheorem{thm}{Theorem}

\newtheorem{lem}{Lemma}

\newtheorem{defn}{Definition}

\newcommand{\bb}{\mathbf{b}}

\newcommand{\bc}{\mathbf{c}}
\newcommand{\bC}{\mathbf{C}}
\newcommand{\bd}{\mathbf{d}}
\newcommand{\bD}{\mathbf{D}}

\newcommand{\bI}{\mathbf{I}}

\newcommand{\bU}{\mathbf{U}}

\newcommand{\bW}{\mathbf{W}}
\newcommand{\bx}{\mathbf{x}}
\newcommand{\bX}{\mathbf{X}}

\newcommand{\calS}{\mathcal{S}}

\usepackage[linesnumbered, ruled]{algorithm2e}

\def\BibTeX{{\rm B\kern-.05em{\sc i\kern-.025em b}\kern-.08em
    T\kern-.1667em\lower.7ex\hbox{E}\kern-.125emX}}
\begin{document}

\title{Provable Data Clustering via Innovation Search}

\author{\IEEEauthorblockN
{\bf Weiwei Li,\ Mostafa Rahmani,\  Ping Li\vspace{0.08in}}
\IEEEauthorblockA{Cognitive Computing Lab \\
Baidu Research\\
10900 NE 8th St. Bellevue, WA 98004, USA \\
\{weiweili.vilock, rahmani.sut,\ pingli98\}@gmail.com}
}

\maketitle

\begin{abstract}
This paper studies the subspace clustering problem in which data points collected from high-dimensional ambient space lie in a union of linear subspaces. Subspace clustering  becomes challenging when the dimension of intersection between  subspaces is large and most of the self-representation based methods are sensitive to the intersection between the span of clusters. In sharp contrast to the self-representation based methods, a recently proposed clustering method termed Innovation Pursuit, computed a set of optimal directions (directions of innovation) to build the adjacency matrix. This paper focuses on the Innovation Pursuit Algorithm
to shed light on its impressive performance when the subspaces are heavily intersected. It is shown that in contrast to most of the existing methods which require the subspaces to be sufficiently incoherent with each other, Innovation Pursuit only requires the innovative components of the subspaces to be sufficiently incoherent with each other. These new sufficient conditions allow the clusters to be strongly close to each other. Motivated by the presented theoretical analysis, a simple yet effective projection based technique is proposed which we show with both numerical and theoretical results that it can boost the  performance of Innovation Pursuit. \\
\end{abstract}


\section{Introduction}
Subspace clustering aims to approximate the data points collected from a high-dimensional space as the union of several low-dimensional linear subspaces. This problem has gained increasing attention in machine learning community due to its wide applications in (e.g.,) motion segmentation~\cite{tron2007benchmark,elhamifar2009sparse,Proc:Yuan_ECCV14}, face clustering~\cite{zhou2014video, vidal2014low, you2016oracle,you2016scalable}, handwritten digits clustering~\cite{you2016divide, ji2017deep, JMLR:v22:18-780}, and manifold learning~\cite{ding2014subspace,wang2016manifold}. At a high level, subspace clustering conducts dimension reduction and clustering simultaneously. Unlike traditional dimension reduction methods~\cite{belkin2003laplacian,jolliffe2016principal}, it is more flexible by allowing more than one  linear structure embedded in the ambient space. In addition, unlike many other clustering algorithms~\cite{park2009simple, jain2010data}, subspace clustering makes use of the linear structure of the data clusters which might lead to superior clustering accuracy.

\subsection{Previous Work}
Most subspace clustering algorithms are composed of  two main stages. In the first step, an affinity matrix is constructed through some measure of similarity calculated for each pair of data points. In the second stage, the spectral clustering algorithm~\cite{ng2002spectral} is applied to the computed  affinity matrix to identify the clusters. The spectral clustering algorithm is guaranteed to deliver exact clustering results~\cite{von2007tutorial} if the graph induced by affinity matrix satisfies the following two properties: the affinity matrix is error less and nodes within each cluster are connected. The subspace clustering algorithms mainly focus on computing an error-less adjacency matrix, i.e.,
the computed affinity value between all pairs of data points which lie in different clusters are zero.

A widely used approach for computing the affinity matrix leverages the linear structure of  clusters and uses the self-representation technique to find the neighbouring points of each data points and compute their corresponding affinity values~\cite{elhamifar2009sparse,liu2012robust,Proc:Liu_NIPS14}.
 For instance, the Sparse Subspace Clustering (SSC) method~\cite{elhamifar2009sparse} computes the affinity values between data column $\mathbf{d}_i$ and the rest of the data points as
\begin{equation}
    \begin{aligned}
        \min_{\boldsymbol{\beta}} \frac{1}{2} \| \mathbf{d}_i - \mathbf{D}\boldsymbol{\beta} \|_2^2 + \lambda \| \boldsymbol{\beta} \|_1 \quad s.t. \quad \mathbf{\beta(i)} = 0,
    \end{aligned}
    \label{eq:SSCC}
\end{equation}
where $\bd_i$ is the  $i^{th}$ column of $\bD$ and $\mathbf{\beta(i)}$ is the $i^{th}$ element of $\mathbf{\beta}$.
Each data point  in a linear cluster can be constructed using a few other data points which lie in the same cluster.
The optimization problem in \eqref{eq:SSCC} utilizes  $\ell_1$-norm to ensure that the non-zero values of the optimal representation vector correspond to the data points which lie in the same cluster that $\bd_i$ lies in.
\cite{soltanolkotabi2012geometric,wang2016noisy} established the performance guarantees for SSC. In~\cite{you2016oracle} and~\cite{you2016scalable} the $\ell_1$-norm of SSC was replaced by the elastic net objective function and orthogonal matching pursuit, respectively. In~\cite{peng2013scalable} and~\cite{you2016divide} two methods were proposed to further speed up the SSC algorithm.

A different line of approaches computes the similarity between all pairs of data points for building the adjacency matrix. For instance, \cite{heckel2015robust} used the absolute value of inner product to measure the similarity  between data points. However, when the clusters are close to each other (i.e., the principal angles~\cite{knyazev2012principal} between the subspaces are small), the inner product might be no longer  reliable for  estimating the affinity. To tackle this challenging scenario,
\cite{rahmani2017subspace,rahmani2017innovation,Proc:Rahmani_NeurIPS19} proposed the Innovation Pursuit (iPursuit) algorithm in which the directions of innovation were utilized to measure the resemblance between the data points. They showed that iPursuit could notably outperform the self-representation based methods especially when the clusters are close to each other. In this paper, we focus on analyzing the theoretical properties for the iPursuit algorithm. Based on our analysis, a simple method is developed to further improve iPursuit when the subspaces are heavily intersected.

\subsection{Contributions}
The iPursuit algorithm has demonstrated superior performance via numerical experiments.
In this paper, we present an analysis of iPursuit under various probabilistic
models. Specifically, we focus on the understanding of  its superior performance  in the challenging cases where subspaces have large dimension of intersection,  whereas other subspace clustering algorithms typically fail to distinguish the clusters. Importantly, our work reveals that the ``innovative component'' of the subspaces is the main factor which affects the  performance of iPursuit. That is, in sharp contrast to many other methods which require the subspaces to be sufficiently incoherent with each other, iPursuit requires the incoherence between the innovative component of the subspaces.

Moreover, we show that the performance of iPursuit can be guaranteed even if the dimension of intersection is in linear order with the rank of the data. To our best knowledge, this is the first time that a subspace clustering algorithm provably tolerates this level of intersection between subspaces.  Inspired by the presented analysis, we introduce a simple yet effective technique to boost the performance of iPursuit. The proposed technique  can reduce the affinities between subspaces and
can enhance the innovation component of the data points which helps the algorithm in distinguishing the clusters.

\subsection{Notation and Definitions}\label{AA}
In this paper, $\bD \in \mathbb{R}^{M \times N}$ is the given data matrix and $\mathbf{d}_i \in \mathbb{R}^{M}$ is its $i$-th column, $\mathbf{D}_{ij}$ is the (i,j)-th entry. The column space of $\mathbf{D}$ is denoted as $\mathit{span}(\mathbf{D})$. The columns of $\mathbf{D}$ lie in a union of $K$ linear subspaces. The value of $K$ might be estimated in various ways, e.g., from the Laplacian matrix~\cite{von2007tutorial}. In this paper, it is assumed  that $K$ is known. Let $\{ \mathcal{S}_k \}_{k=1}^K$ denote the span of the clusters. The sub-matrix of $\bD$ whose columns lie in $\mathcal{S}_k$ is denoted as $\mathbf{D}_k$. Subspace $\mathcal{S}_k$ contains $n_k$ data points and the dimension of $\mathcal{S}_k$ is $m_k$. The columns of  orthonormal matrix $\mathbf{U}_k \in \mathbb{R}^{M \times m_k}$ form a basis for $\mathcal{S}_k$. To explicitly model the intersection between subspaces, each basis $\bU_k$ is written as
\begin{equation}\label{eq:basisseq}
    \mathbf{U}_k = \left[ \mathbf{U}, \hat{\mathbf{U}}_k \right] \: .
\end{equation}
Here $span(\mathbf{U})$ models the non-trivial intersection among all the subspaces while $\mathbf{\hat{U}}_k$ is the innovation component of $\mathcal{S}_k$. Define $s$ as the dimension of the column space of $\mathbf{U}$ (a large value of $s$ means the subspaces are heavily intersected and they are very close to each other). Note that  the formulation of $\mathbf{U}_k$ in \eqref{eq:basisseq} does not lose  generality and it simplifies the representation of the challenging clustering scenarios.

The $i$-th point in cluster $\mathcal{S}_k$ is denoted as $\mathbf{d}_i^{(k)}$, $\mathbf{d}_i^{(k)} = \mathbf{U}_k \mathbf{b}_i^{(k)}$, and $\|\mathbf{d}_i^{(k)} \|_2 = \|\mathbf{b}_i^{(k)} \|_2$. The direct sum of  subspaces $\calS_1$ and $\calS_2$ is denoted by $\calS_1 \oplus \calS_2$. We define $\mathcal{S}_{-k} = \oplus_{i \neq k} \mathcal{S}_i$  and $\mathbf{U}_{-k}$ as an orthonormal basis matrix for  $\mathcal{S}_{-k}$. Matrix  $\mathbf{D}_{-k}$ denotes matrix $\mathbf{D}$ after removing the columns of $\mathbf{D}_k$. We also define $\mathit{aff}_{\infty}(\mathbf{W}_1,\mathbf{W}_2)$ as the \emph{cosine}  of the smallest principal angle~\cite{knyazev2012principal} between $span(\mathbf{W}_1)$ and $span(\mathbf{W}_2)$. We assume that the columns of $\mathbf{D}$ are normalized to  unit $\ell_2$-norm.

\section{Analysis of the iPursuit Method}\label{sec:thm}
Here we present a theoretical analysis of iPursuit method.

\subsection{The iPursuit Algorithm}
The iPursuit algorithm~\cite{rahmani2017subspace,rahmani2017innovation,Proc:Rahmani_NeurIPS19} computes an optimal direction for each data point and the optimal direction is used to measure the similarity between that data point and the rest of the data. The optimal direction, termed direction of innovation corresponding to $\bd_i$, is computed as the optimal point of
 \begin{equation}
\begin{aligned}
    \min \quad \| \mathbf{c}^T \mathbf{D} \|_1 \quad s.t. \quad \mathbf{c}^T \mathbf{\bd_i}=1.
\end{aligned}
\label{eq:opt_mainn}
\end{equation}
The optimization problem (\ref{eq:opt_mainn}) finds an optimal direction corresponding to $\bd_i$ which has a non-zero inner product with $\bd_i$ and it has the minimum projection on the rest of the data. Since the $\ell_1$-norm is used in the cost function, (\ref{eq:opt_mainn}) prefers a direction which is orthogonal to a maximum number of data points. Suppose $\bd_i$ lies in $\calS_k$ and assume that $\calS_k \notin \calS_{-k}$ and define $\calS_k^{\perp}$ as the span of $(\bI - \bU_{-k} \bU_{-k}^T) \bU_k$ (subspace $\calS_k^{\perp}$ is the innovative component of $\calS_k$ which does not lie in the direct sum of other subspaces). If some sufficient conditions are satisfied,  the optimal solution of (\ref{eq:opt_mainn}) lies in $\calS_k^{\perp}$ which means that it  is orthogonal to all the data points which lie in the other clusters. Accordingly,  vector $|\bc^T \bD|$ can be used as a proper vector of affinity values between $\bd_i$ and the rest of data points. Algorithm 1 presents the iPursuit algorithm which utilizes spectral clustering to process the computed adjacency matrix. The optimization problem in Algorithm 1 is equivalent to (\ref{eq:opt_mainn}) but it computes all the directions corresponding for all the data points simultaneously.

\begin{algorithm}
\caption{Innovation Pursuit with Spectral Clustering~\cite{rahmani2017subspace,rahmani2017innovation,Proc:Rahmani_NeurIPS19}}

\smallbreak
\textbf{Input:} Data matrix $\mathbf{D} \in \mathbb{R}^{M \times N}$.

\smallbreak
{Define $\bC^{*} \in \mathbb{R}^{M \times N}$} as the optimal solution of
\begin{eqnarray}\notag
\begin{aligned}
{\min}\:   \| \bC^T \bD \|_{1}  \quad \text{subject to} \quad  \text{diag}( \bC^T \bD) = \textbf{1} \:.
\end{aligned}
\label{eq:kolli}
\end{eqnarray}

\smallbreak
 Sparsify the adjacency matrix $\bW =  |{\bC^{*}}^T \bD|$ via keeping few dominant non-zero elements of each row.

\smallbreak
 Apply spectral clustering to $\bW = \bW + \bW^T$.

\smallbreak
\textbf{Output} The identified clusters by the spectral clustering algorithm.
\end{algorithm}

\subsection{Data Model and Innovation Assumption}
Define $\bc_i^{*}$ as the optimal point of (\ref{eq:opt_mainn}) (the optimal direction corresponding to $\bd_i$).
In the presented results, we establish the sufficient conditions to guarantee that all the computed directions
yield proper affinity vectors.
A proper affinity vector is defined as follows.

\begin{defn}
Suppose $\bd_i$ lies in $\calS_k$. The optimal direction $\mathbf{c}_i^{*}$ yields a proper affinity vector if
all the data points corresponding to
 the nonzero entries of  ${\mathbf{c}_i^{*} }^T \mathbf{D}$ lie in $\calS_k$.
\end{defn}
If all the directions yield proper affinity vectors, then iPursuit  delivers an error-less affinity matrix.   Next, we  provide the sufficient conditions for a deterministic data model, which reveal the requirements of the algorithm about the distribution of the subspaces and the distribution of the data points inside the subspaces. Then, we present performance guarantees with  random models for the distribution of data points/subspaces. The presumed models can be summarized as follows.
\begin{enumerate}
    \item \emph{Deterministic Model}: both $\{ \mathbf{U}_k \}_{k=1}^{K}$ and $\{ \mathbf{b}_i^{(k)} \}_{i=1}^{n_k}$ are deterministic.
    \item \emph{Semi-Random Model}: the subspaces are deterministic but $\{ \mathbf{b}_i^{(k)} \}_{i=1}^{n}$ are sampled uniformly at random from $\mathbb{S}^{m-1}$ where $\bb_i^{(k)}$ was defined as $\bb_i^{(k)} = \bU_k^T \bd_i^{(k)}$.
    \item \emph{Fully Random Model}: the subspaces $\{ \mathcal{S}_k \}_{k=1}^{K}$ are  sampled uniformly at random from Grassmannian manifold with dimension $m$ and $\{ \mathbf{b}_i^{(k)} \}_{i=1}^{n}$ are sampled uniformly at random from $\mathbb{S}^{m-1}$.
\end{enumerate}
 Note that in the fully random model, we assume $\{ \mathcal{S}_k \}_{k=1}^K$ have the same dimension $m$ and same number of points $n$. In the first three theorems, it is assumed that each cluster carries an innovation with respect to the other clusters. Specifically, it is assumed that each subspace satisfies the Innovation Assumption described as follows.
\begin{defn}
The Innovation Assumption  is satisfied if for any $k=1,...,K$:
\begin{equation*}
    \mathcal{S}_k \notin \oplus_{i \neq k} \mathcal{S}_i.
\end{equation*}
\end{defn}
Note that Innovation Assumption does not mean independence between the subspaces and it allows the subspaces to have large dimension of intersection. In the rest of the paper, by the innovation subspace corresponding to $\calS_k$, we mean the projection of $\calS_k$ onto the complement of $\oplus_{i \neq k} \mathcal{S}_i$.
Define $\mathbf{\hat{U}}_{-k}$ as an orthonormal basis of $\oplus_{i \neq k}\mathit{span}(\mathbf{\hat{U}}_i)$. Accordingly, Innovation Assumption is equivalent to $\mathit{span}(\mathbf{\hat{U}}_k) \notin \mathit{span}(\mathbf{\hat{U}}_{-k})$. The presented results show that iPursuit yields an error-less affinity matrix even if $s$ (the dimension of intersection between the span of clusters) is large.

Two main factors play an important role in  the performance of the subspace clustering algorithms: the affinity between each pair of subspaces and the distribution of the data points inside the subspaces.  The following definition specifies these parameters which are used in the presented results.
\begin{defn}\label{defn:metric}
Define $h_1$ and $h_2$ as
    \begin{equation*}
        h_1 = \min_{1 \leq k \leq K \atop \delta \in \calS_k} \inf_{\|\delta\|_2 = 1} \|\delta^T \mathbf{D}_k \|_1, \quad      h_2 = \max_{1 \leq k \leq K \atop \delta \in \calS_k} \sup_{\|\delta\|_2 = 1} \|\delta^T \mathbf{D}_k \|_1 \:.
    \end{equation*}
Define $\{t_i\}_{i=1}^3$ as the minimum positive real numbers such~that
\begin{eqnarray}
\begin{aligned}
      &\max_{1 \leq k \leq K} \mathit{aff}_{\infty}(\mathbf{\hat{U}}_i,\mathbf{\hat{U}}_{-i}) \leq t_1, \:\\
      &\max_{1 \leq i \neq j \leq K} \mathit{aff}_{\infty}(\mathbf{\hat{U}}_i,\mathbf{\hat{U}}_j) \leq t_2, \: \\
      &\max_{1 \leq k \leq K} \max_{1 \leq i \leq n_k} \mathit{aff}_{\infty}(\mathbf{d}_i^{(k)},\mathbf{U}_{-k}) \leq t_3 \:.
\end{aligned}
     \label{eq:aff_inf}
\end{eqnarray}
\end{defn}
Quantity $h_1$ is known as the permeance statistics~\cite{lerman2015robust} which characterises how well the data points are distributed in each subspace. A large value of $h_1$   means that data points are well scattered in their corresponding subspaces. In other words, $\frac{h_2}{h_1}$ is large if the data points are concentrated along certain directions in the subspace and vice versa.
 Recall $\mathit{aff}_{\infty}(\mathbf{W}_i,\mathbf{W}_{j})$ is the \emph{cosine} value of the smallest principal angle between $span(\mathbf{W}_i)$ and $span(\mathbf{W}_j)$. The parameters $t_1$ and $t_2$ determine the coherence between the innovative components of the subspaces. The smaller $t_1$ and $t_2$ are, the more distinguishable the innovative components are.

 Another measure of affinity between subspaces is
\begin{equation*}
    \mathit{aff}(\mathcal{S}_k,\mathcal{S}_l) = \sqrt{\frac{\cos^2{\theta_1}+...+\cos^2{\theta_{\min(m_k,m_l)}}}{\min(m_k,m_l)}},
\end{equation*}
where $\{ \theta_i \}_{i=1}^{\min(m_k,m_l)}$ are the principal angles between $\mathcal{S}_k$ and $\mathcal{S}_l$~\cite{soltanolkotabi2012geometric}. A key difference between $\mathit{aff}(\mathcal{S}_k,\mathcal{S}_l)$ and $\{ t_i \}_{i=1,2}$ is, $\mathit{aff}(\mathcal{S}_k,\mathcal{S}_l)$ takes into account the intersection between the subspaces, hence $\mathit{aff}(\mathcal{S}_k,\mathcal{S}_l)$ will be nearly equal to one when $s$ is large, while $\{ t_i \}_{i=1,2}$ only consider the similarity between innovation parts.
The quantity $t_3$
characterises the innovative component of the data points, i.e., a small value of $t_3$ means that  data points are  close to their corresponding innovation subspaces. In order to understand the importance of $t_3$, consider a scenario where some of the data points completely lie in  $span(\mathbf{U})$. In this scenario, no clustering algorithm can assign a clustering label to those data points because they could belong to all the clusters (since they completely lie in the intersection subspace).

\subsection{Theoretical Guarantees With the Deterministic Model}\label{sec:deterministic}
Theorem~\ref{thm:preserve-deterministic} provides sufficient conditions to guarantee that iPursuit yields an error-less affinity matrix.
\begin{thm}\label{thm:preserve-deterministic}
 Assume Innovation Assumption is satisfied and
\begin{eqnarray}
\begin{aligned}
h_1 \sqrt{ 1-(K-2)t_2 } \geq h_2 (\sqrt{\frac{t_3^2}{1-t_3^2}}+ t_1), \\ h_1 \sqrt{K-1} \geq h_2 (\sqrt{\frac{t_3^2}{1-t_3^2}}+ 1) \: ,
    \end{aligned}\label{eq:condition1}
\end{eqnarray}
then    all the optimal directions yield proper affinity vectors and
    iPursuit constructs an error-less affinity matrix.
\end{thm}
The sufficient conditions are  satisfied if $\{ t_i \}_{i=1}^3$ are sufficiently small and $\frac{h_1}{h_2}$ is close enough to $1$. Geometrically, it means that  the algorithm requires the innovative component of each pair of subspaces (measured by $\{ t_i \}_{i=1}^3$) are sufficiently away from each other and the data points are well scattered inside the subspaces (measured by $\frac{h_1}{h_2})$.
 Importantly, these conditions allow the subspaces to have a large dimension of intersection because in contrast to the other methods, iPursuit requires incoherence between the innovative components not the subspaces themselves. Consider an extreme case where $t_1 = t_2 =0$. The first inequality requires $\frac{t_3^2}{1-t_3^2} < 1$ hence $t_3 < \frac{\sqrt{2}}{2}$. Thus, if the data points are distributed well in the subspace,  $t_3$ can be almost equal to $\frac{\sqrt{2}}{2}$ which means that the dimension of intersection between the subspaces might be close to $\frac{m}{2}$ or at least linear with $m$. To the best of our knowledge, no existing subspace clustering algorithms can provably allow this degree of intersections. 

\subsection{Theoretical Guarantees with the Semi-Random Model}
 In the semi-random model, it was assumed that the data points in each subspace are  distributed uniformly at random. 
\begin{thm}\label{thm:preserve-semi}
Suppose the data follows the Semi-Random Model and the Innovation Assumption is satisfied. Define
\begin{eqnarray*}
\begin{aligned}
    & h_1 = \sqrt{\frac{2}{\pi}} \frac{n}{\sqrt{m}} - 2 \sqrt{n} - \sqrt{\frac{2n \log n}{m-1}}
     \\
     &h_2 = \frac{n}{\sqrt{m}} + 2\sqrt{n} + \sqrt{\frac{2n \log n}{m-1}} \:,
    \end{aligned}
\end{eqnarray*}
Define  $\{ t_i \}_{i=1}^3$ as in Definition~\ref{defn:metric}, and
\begin{eqnarray}
\begin{aligned}
& \zeta = \frac{(m-s)\left[ t_3^2-(K-1)t_2^2 \right]}{s(1-t_3^2)} \:.
\end{aligned}
\label{eq:new_int2}
\end{eqnarray}
If (\ref{eq:condition1}) is satisfied and $\zeta >1$, then all the optimal directions yield proper affinity vectors and iPursuit constructs an error-less affinity matrix with probability at least
$
    1-\frac{2}{n} - 2Ne^{-\epsilon^2}
$
where $\epsilon = \frac{-(\sqrt{s}+\frac{\zeta s}{\sqrt{m-s}})}{2} + \frac{\sqrt{(\sqrt{s}+\frac{\zeta s}{\sqrt{m-s}})^2+2s(\zeta-1)}}{2} $.
\end{thm}

To understand the significance of the result, consider an extreme scenario in which  $t_1 = t_2 = 0$ (i.e., the innovative components are fully incoherent with each other).  If there are a sufficient number of data points in each cluster (i.e., $n/\sqrt{m}$ is large), we have $\frac{h_1}{h_2} \approx \sqrt{\frac{2}{\pi}}$, hence we need $t_3 < \sqrt{\frac{2}{2+\pi}}$. {To guarantee that $\zeta > 1$,  $s$ (the dimension of intersection) can not be larger than $\frac{2m}{2+\pi}$. This means that iPursuit is  able to allow $s$ to be in linear order with $m$  while other algorithms such as SSC and TSC need $\max_{k,l} \mathit{aff}(\mathcal{S}_k,\mathcal{S}_l)$ to be in the order of $O(\frac{1}{\log N})$~\cite{wang2016noisy, soltanolkotabi2012geometric, heckel2015robust}. This assumption will be violated if $s$ is in linear order with $m$. Therefore, our results require a much more weaker condition on the affinities between these subspaces. }
The new information we learn from this theorem is that if the dimension of the subspaces increases, more data  points is required to guarantee the performance of iPursuit. This requirement is intuitively correct because as $m$ increases, more number of data points is required to populate the clusters.

\subsection{Theoretical guarantees with the fully-random model}\label{sec:limiting-principal}

We now consider the fully random model. In order to model the intersection between subspaces explicitly, assume $\mathit{span}(\mathbf{U})$ and $\{\mathit{span}(\mathbf{\hat{U}}_k)\}_{k=1}^K$ are sampled uniformly at  random from Grassmannian manifolds with dimension $s$ and $(m-s)$, respectively. Note that this setting is different from the traditional fully random model, in which $\{\mathit{span}(\mathbf{U}_k)\}_{k=1}^K$ are sampled independently from a Grassmannian manifold with dimension $m$. In this section, it is assumed that $s$ is linear with $m$. In order to derive the sufficient conditions, first we need to establish  upper-bounds for $t_1$ and $t_2$ in \eqref{eq:aff_inf}. Under the fully random model, $\mathit{aff}^2_{\infty}(\mathbf{\hat{U}}_i,\mathbf{\hat{U}}_{-i})$ is the largest root of a multivariate beta distribution with the following limiting behavior as $m \rightarrow \infty$~\cite{johnstone2008multivariate}
\begin{eqnarray}
\label{eq:cos-principal}
\begin{aligned}
 &   \mathit{aff}^2_{\infty}(\mathbf{\hat{U}}_k,\mathbf{\hat{U}}_{-k}) \leq T_{M,m,K,s} +\sigma |F_1|,
\end{aligned}
\end{eqnarray}
where \vspace{-0.15in}

{\scriptsize
\begin{align}
T_{M,m,K,s} =     \frac{(K-1)(m-s)+m+2\sqrt{(m-s)\left[ (K-2)(m-s)+m \right]}}{M}
\end{align}
}\vspace{-0.1in}

\noindent and  $F_1$ is a special random variable which has the mean $\sqrt{2(K-1)(m-s)+s}$, with typical fluctuations of $O(m^{1/6})$ and atypical large deviations of $O(m^{1/2})$~\cite{nadal2011simple} . This means random variable $F_1$ is in the order of $O(m^{1/2})$ almost surely. It addition, $\sigma$ scales with $O(\frac{m^{1/3}}{M})$ and the first term in RHS of \eqref{eq:cos-principal} has an order of $O(\frac{m}{M})$, hence for large $m$ the RHS of \eqref{eq:cos-principal} is dominated by its first term almost surely. Similarly $\mathit{aff}^2_{\infty}{(\mathbf{\hat{U}}_i,\mathbf{\hat{U}}_j)} \approx \frac{4(m-s)}{M}$. 

\begin{thm}\label{thm:preserve-fully}
Define $\zeta$ and $\epsilon$ as in \eqref{eq:new_int2}. Assume conditions in~\cite{johnstone2008multivariate}  and there exist positive constants $\hat{t}_1$, $\hat{t}_2$ and $t_3$ such that if
\begin{align}\notag
  &   t_1^2 := \hat{t}_1\cdot T_{M,m,K,s}, \hspace{0.3in}
     t_2^2 :=  \hat{t}_2 \cdot \frac{4(m-s)}{M},
\end{align}
we have $\zeta > 1$, and both inequalities in \eqref{eq:aff_inf} are satisfied. Then all the optimal directions yield proper affinity vectors and iPursuit builds an error-less affinity matrix with probability at least
$
    1-\frac{2}{n} - 2Ne^{-\epsilon^2}  .
$
\end{thm}
The sufficient conditions in Theorem~\ref{thm:preserve-fully} roughly state that we need $\frac{(K-1)m}{M}$ to be sufficiently  small and this confirms our intuition since it  implies that it is more likely to sample subspaces with small affinities.

\section{Enhancing the Performance of  iPursuit}\label{sec:thresholding}

In this section, we address a performance bottleneck of iPursuit. While iPursuit performs well in handling the subspaces with high dimension of intersection, the algorithm could fail to obtain a proper affinity vector for $\bd_i$ when the projection of $\bd_i$ in the span of $\bU$ (the intersection part) is strong. The main reason is that if $\bd_i$ is close to the span of $\bU$, then $\bc_i^{*}$ (the optimal point of (\ref{eq:opt_mainn})) should have a large $\ell_2$-norm to lie in the innovation subspace and to satisfy the constraint because $\bd_i$ is strongly aligned with the span of $\bU$. Therefore, the optimal point of (\ref{eq:opt_mainn}) may not lie in the innovation subspace or be close to it since the cost function of (\ref{eq:opt_mainn}) increases with $\| \bc_i^{*} \|_2$.

We present a technique which enables  iPursuit  to handle data points with  weak projection on their corresponding innovation subspaces.
Define $\bD_U$ as $(\bI - \bU \bU^T)\bD$ and $\bD_{U^{\perp}} = (\bI - \bU^{\perp} {\bU^{\perp}}^T)$ where $\bU^{\perp}$ is an orthonormal basis for the column space of $\bI - \bU \bU^T$. In other words, $\bD_U$ is the projection of the data points in $span(\bU)$ and $\bD_{U^{\perp}}$ is the projection of the data points in the complement of $span(\bU)$, i.e., the columns of $\bD_{U^{\perp}}$ are the projection of  data points in the innovation subspaces. When the dimension of intersection is large, $\| \bD_U \|_F$ is much larger than $\| \bD_{U^{\perp}}\|_F$, i.e., most of the power of the data lies in $span(\bU)$. Accordingly, we approximate $span(\bU)$ with the span of the dominant left singular vectors. The next theorem shows that the first singular vector is close to $span(\bU)$ when $m/(m - s)$ is large.

\begin{thm}\label{thm:nrm-bnd}
{
Assume the semi random data model. Define $\bx_1$ as the first left singular vector of $\bD$, and $\lambda_1$/$\lambda_2$ as $\| \bU^T \bx_1 \|_2$/ $\| {\bU^{\perp}}^T \bx_1 \|_2$. For positive constant  $\kappa \in (\frac{m-s}{m},1)$ we have
\begin{equation}\label{eq:nrm-bnd}
    \frac{\lambda_1}{\lambda_2} \geq \frac{\kappa'-1}{2\sqrt{\kappa'}},
\end{equation}
with probability at least $1-\frac{1}{n}-2Kne^{-\epsilon^2}$, where $$\kappa' =  \frac{m\left[ \sqrt{ \frac{2Kn}{\pi m}} - 2 - \sqrt{\frac{2\log n}{m-1}} \right ]^2 }{\left[1+(K-1)t_2 \right]n(m-s)\kappa}$$ and
\[
\epsilon = \frac{(m-s)B}{A +\sqrt{A^2+2(m-s)B}},
\]
}
where $A = \sqrt{m-s}+\frac{\kappa\sqrt{s}}{1-\kappa}$ and $B =\frac{\kappa s }{(1-\kappa)(m-s)}-1 $.
\end{thm}
The probability bound in Theorem~\ref{thm:nrm-bnd} is approximately equal to  $1$ when $n$ is sufficiently large and $m > 3\log N$. With large $n$, the order of RHS of \eqref{eq:nrm-bnd} is roughly equal to $O\left(\sqrt{\frac{K\:m}{\left[ 1+(K-1)t_2 \right]\kappa (m-s)}}\right)$, which means that $\| \bU^T \bx_1 \|_2$ increases when $m/(m-s)$ increases.

Motivated by the observation that the intersection subspace is close to the span of dominant singular vectors, we improve the performance of iPursuit by  filtering the projection of the data points  in the span of the dominant singular vectors. 
Define $\bX$ as the matrix of left singular vectors corresponding to the non-zero singular values and  $\bX_{\hat{s}:}$ is formed by removing the first $\hat{s}$ columns of $\bX$. The parameter $\hat{s}$ is our estimate of $s$. Accordingly, we update each column of $\bD$ as $\bd_i \Leftarrow \frac{\bX_{\hat{s}:} \bX_{\hat{s}:}^T \bd_i}{\| \bX_{\hat{s}:} \bX_{\hat{s}:}^T \bd_i \|_2}$ to remove the projection of each data point in the span of the dominant singular vectors and enhance the innovative component of the data points.

\section{Numerical Experiments}\label{sec:numerical}
We report two sets of experiments. First, we analyze the approximation studied in \eqref{eq:cos-principal}; second, we show the superior performance of iPursuit equipped with  the technique presented in Section~\ref{sec:thresholding}, using both synthetic and real datasets, measured by clustering accuracy. We refer to iPursuit method equipped with the  technique  in Section~\ref{sec:thresholding} as Enhanced iPursuit.

\begin{figure}[t!]
 \centering
    \includegraphics[width=0.30\textwidth]{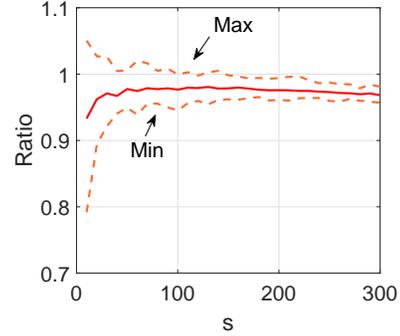}
    \vspace{-0.15in}
    \caption{ We fix $M=10000$, $K=10$ and change $s$ from $10$ to $300$, $m$ is set to be $1.5s$. For each set of parameters we simulate $50$ pairs of random subspaces with dimension $(m-s)$ and $\left [(K-1)(m-s)+s\right ]$, respectively, and calculate the \emph{cosine} values of their smallest principal angle (denoted here as $\cos{\theta_1}$). The ``ratio'' in the figure is defined as the ratio between $\cos{\theta_1}$ and the estimated quantity in \eqref{eq:cos-principal}. We plot the mean values, together with the min and max values (dahsed), of the ratios generated in the simulation.  }
\label{fig:approximation}
\end{figure}

In Figure~\ref{fig:approximation}, we simulated pairs of random subspaces under different settings and calculated the ratio between true \emph{cosine} values and the corresponding estimations (i.e., first term in RHS of \eqref{eq:cos-principal}).  As we increase $s$, the approximation tends to be very stable and slightly below $1$ which confirms our analysis in Section~\ref{sec:limiting-principal}. In this simulation we fix the ratio between $m$ and $s$. With fixed $M$, the first term at the RHS of \eqref{eq:cos-principal} is more likely to be the dominant term as we increase $m$.

\subsection{Results on Synthetic Data}
We first compare the performances of iPursuit with Enhanced iPursuit using synthetic datasets. Specifically, two groups of experiments are considered: in the first group, we fix $M=60$, $m=s+2$, $K=10$, and increase $s$ from $10$ to $40$; in the second group, we fix $M=60$, $s=40$, $m=42$, and increase $K$ from $5$ to $10$. In both groups we choose $\hat{s} = s-5$.

\begin{figure}[h!]
\begin{center}
\mbox{
\includegraphics[width=1.8in]{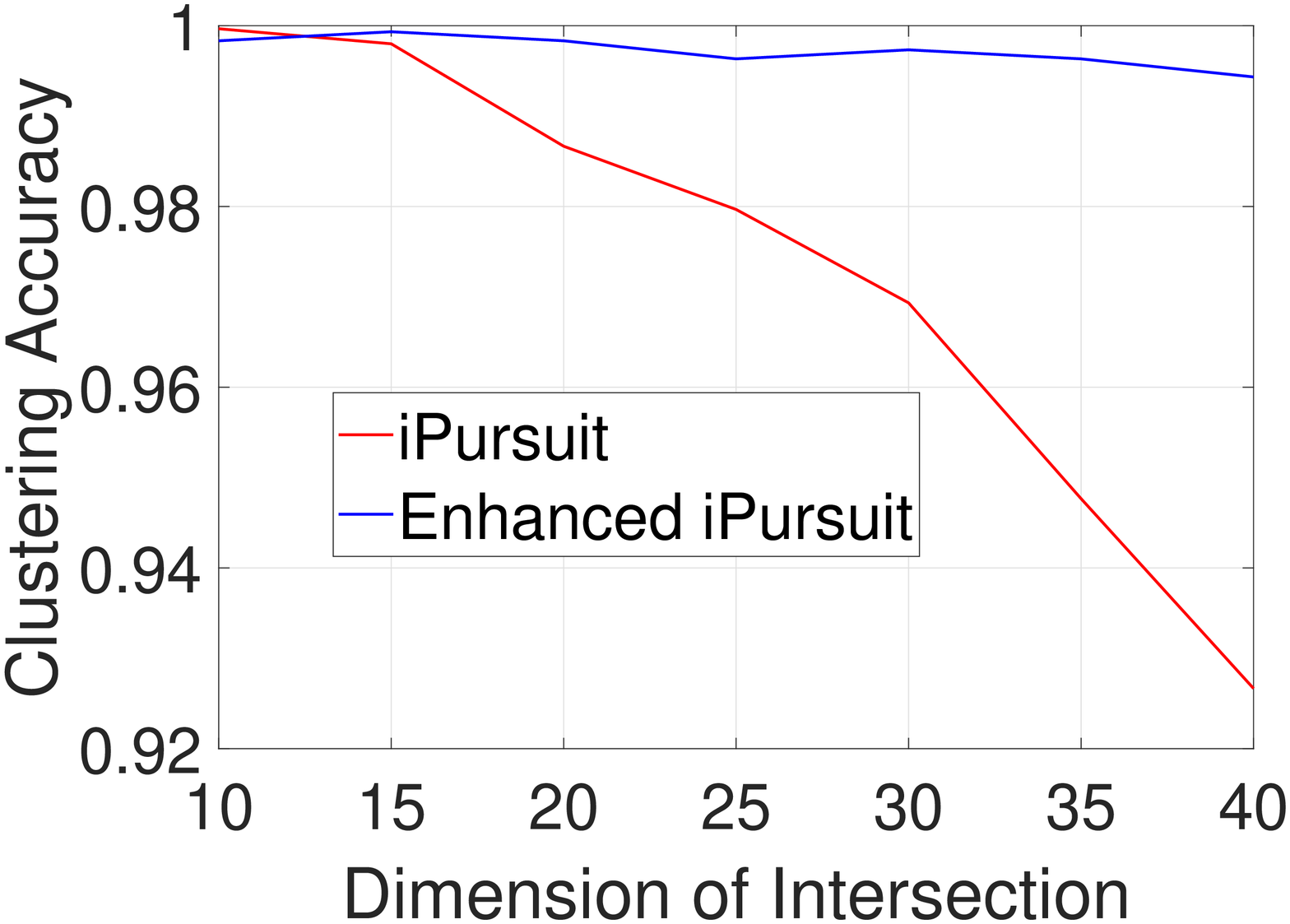}\hspace{-0.15in}
\includegraphics[width=1.8in]{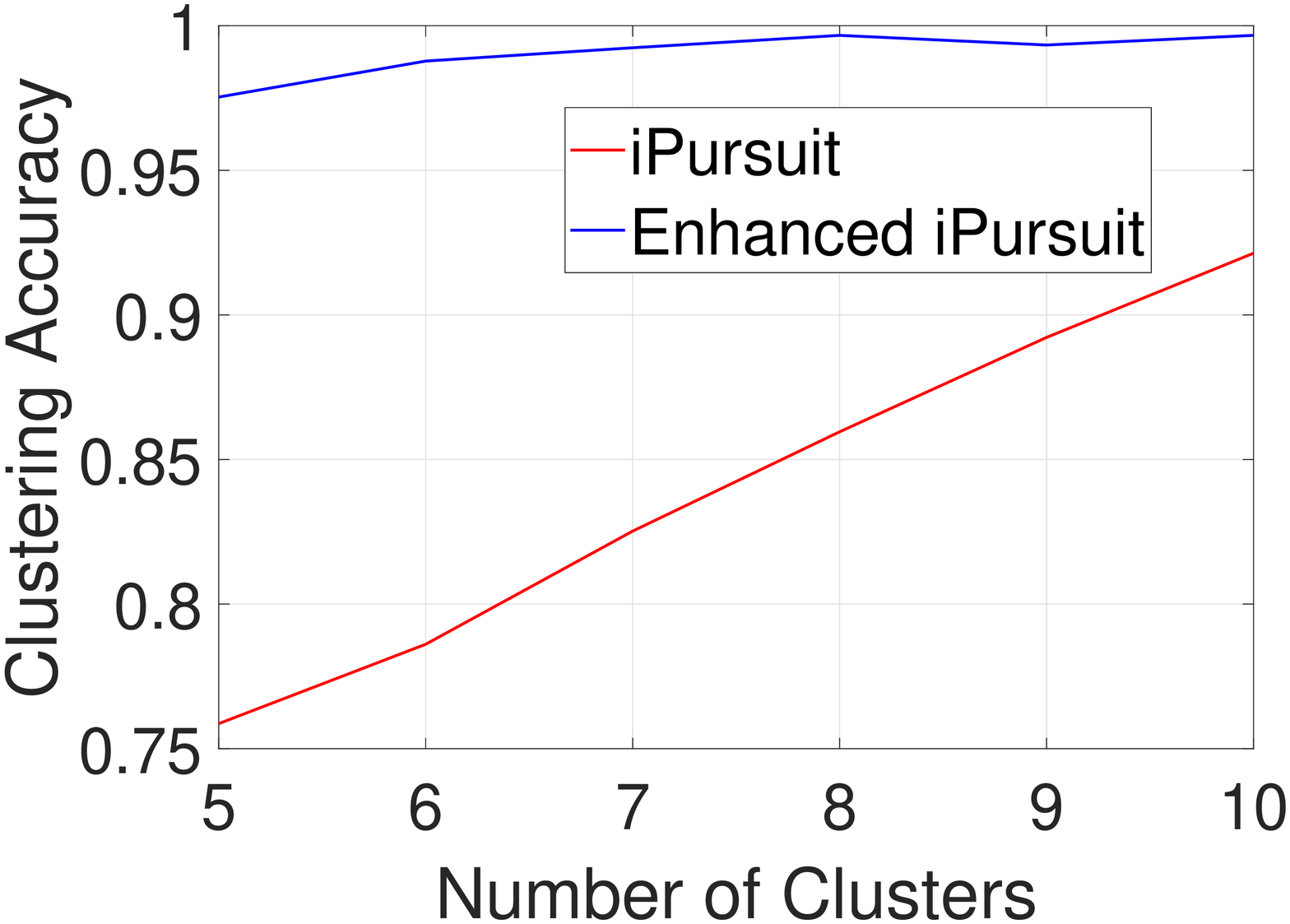}
}\end{center}

\vspace{-0.15in}

\caption{Clustering accuracy on the synthetic datasets.}\label{fig:syn}\vspace{-0.1in}
\end{figure}

Figure~\ref{fig:syn} illustrates that Enhanced iPursuit is   outperforming iPursuit. In the first plot, the clustering accuracy decreases as the dimension of intersections $s$ increases. In the second group, we observe that clustering accuracy of iPursuit increases with $K$ which confirms our analysis in Section~\ref{sec:thm}.

\begin{table*}[t]
\begin{center}
\caption{Clustering accuracy on MNIST and ZIPCODE}\label{tab:mnist}
\begin{tabular}{@{}ccccccccc@{}}\hline
   &      & Enhanced iPursuit & iPursuit  & SSC & TSC & OMP & ENSC & KMEANS \\ \hline
   &MNIST  & \textbf{97.48\%} & 89.23 \% & 72.92\% & 81.31\% & 78.73\% & 81.79\% &65.15\% \\ \hline
   &ZIPCODE &\textbf{72.12\%} &66.25\% &45.49\% & 66.31\% & 17.6\% & 45\% &67.06\% \\ \hline
\end{tabular}\vspace{-0.1in}
\end{center}
\end{table*}

\begin{figure}[b!]
\vspace{-0.15in}

\mbox{
\includegraphics[width=1.7in]{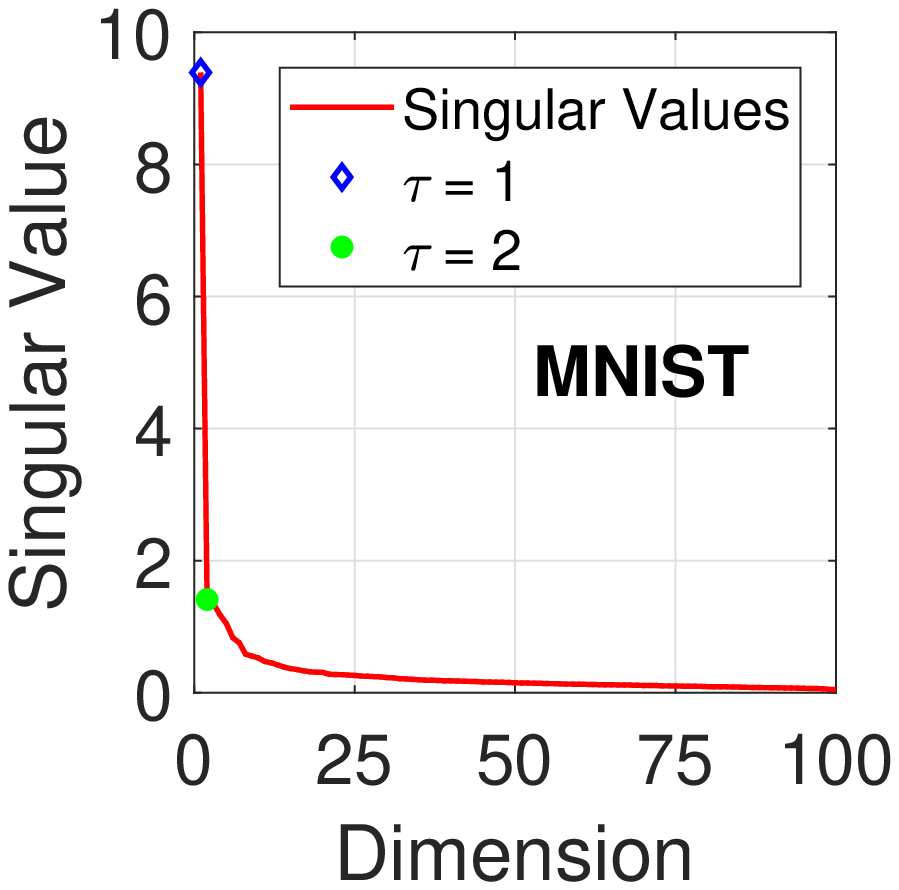}
\includegraphics[width=1.7in]{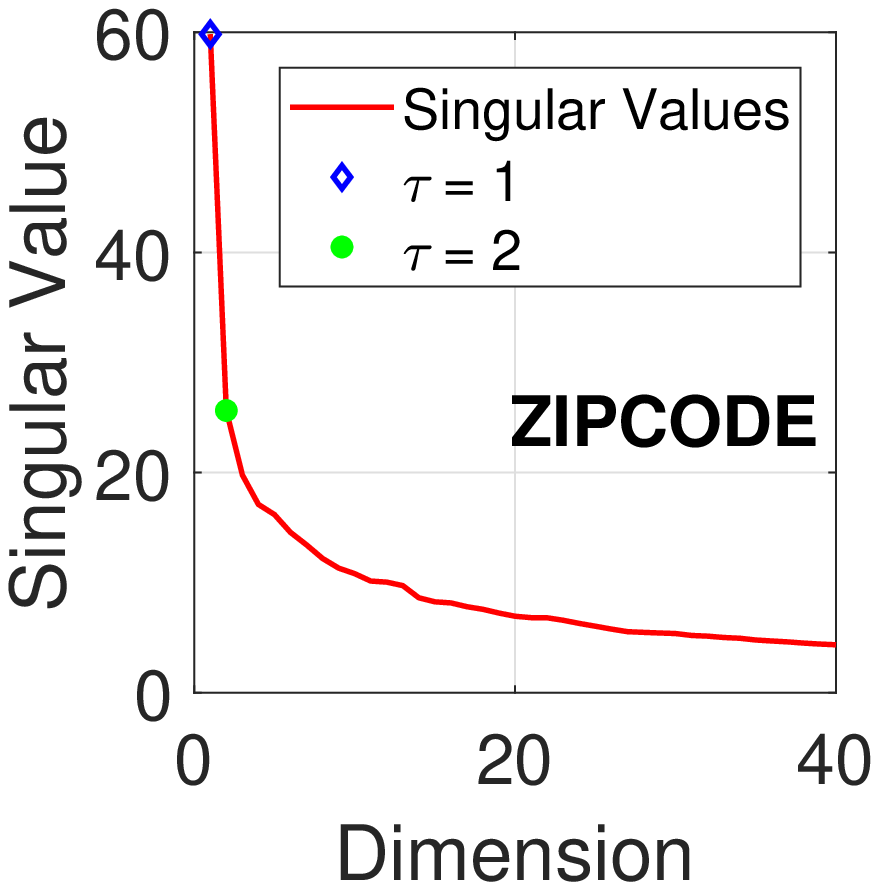}
}
\vspace{-0.2in}
\caption{Plots of the singular values for MNIST (left) and ZIPCODE (right).}\label{fig:singular}
\end{figure}

\subsection{Results on Handwritten Digits}
We now demonstrate the results of Enhanced iPursuit on two popular handwritten digits datasets: MNIST and ZIPCODE. For MNIST, we use the CNN extracted features which are 3472-dimensional vectors. Using PCA, we  reduced the dimension to $500$~\cite{you2016scalable}. For ZIPCODE, we conducted a PCA step to reduce the dimension from $256$ to $40$. In Figure~\ref{fig:singular}, we observe  a clear gap between the first and second singular values in both MNIST and ZIPCODE; hence we set $\hat{s} = 1$. The results are demonstrated in Table~\ref{tab:mnist} (for both experiments we fix $\tau = 1$.). One can observe that the proposed technique enhanced the performance of iPursuit from $80\%$ to nearly $98\%$ on MNIST and $66\%$ to $72\%$ on ZIPCODE. In both  datasets, Enhanced iPursuit   yielded the best clustering accuracy.

\section{Conclusion}
While many subspace clustering algorithms cannot deal with heavily intersected subspaces, iPursuit~\cite{rahmani2017subspace,rahmani2017innovation,Proc:Rahmani_NeurIPS19} was shown to notably outperform the other methods in this challenging setting. In this paper, we provided an analysis of iPursuit to understand its capability in handling heavily intersected subspaces.
It was shown that in sharp contrast to the other methods which require the subspaces to be sufficiently incoherence with each other, iPursuit requires  incoherence between the innovative components of the subspaces. In addition, we proved that iPursuit can yield provable performance even if the dimension of intersection is in linear order with rank of the data.
We also proposed a promising projection-based method to further improve iPursuit.

\bibliographystyle{plain}
\bibliography{reference_ip}

\newpage

\onecolumn
\section*{Proofs of Theorems and Auxiliary Lemmas}

\vspace{0.2in}

\begin{lem}\label{lem:prj}
Let $\mathbf{P}_1,...,\mathbf{P}_n$ be $n$ orthonormal matrices in $\mathbb{R}^{r_1 \times r_2}$ and assume
\begin{equation*}
    \mathit{aff}_{\infty}{(\mathbf{P}_i,\mathbf{P}_j)} \leq a \quad \textit{for any $1 \leq i\neq j \leq n$,}
\end{equation*}
here $a \leq \frac{1}{2}$ is a constant. Then for any vector $\mathbf{g} \in \oplus_{i=1}^n span(\mathbf{P}_i)$ with unit $\ell_2$-norm, the following two inequalities hold:
\begin{align*}
   \inf \sum_{i=1}^n \| \mathbf{P}_i^T \mathbf{g} \|_2^2 \geq 1 - (n-1) a,\quad \sup \sum_{i=1}^n \| \mathbf{P}_i^T \mathbf{g} \|_2^2 \leq 1 + (n-1) a.
\end{align*}
\end{lem}

\vspace{0.2in}

\noindent
\textbf{Proof of Lemma~\ref{lem:prj}}:
We will prove the first part of Lemma~\ref{lem:prj} by using mathematical induction on $n$, the second part can be proved similarly. For $n=1$, the result is straightforward. Now consider for $n=2$, we write $\mathbf{g}= \left [  \mathbf{P}_1 \mathbf{P}_2  \right ] \begin{bmatrix}
           \mathbf{g}_1 \\
           \mathbf{g}_2
         \end{bmatrix}$, here both $\mathbf{g}_1$ and $\mathbf{g}_2$ are vectors in $\mathbb{R}^{r_2}$.
Note that $\| \mathbf{g} \|_2 = 1$ has unit norm, we then have
\begin{equation*}
     \left [ \mathbf{g}_1^T \mathbf{g}_2^T  \right ] \begin{bmatrix}
           \mathbf{P}_1^T \\
           \mathbf{P}_2^T
         \end{bmatrix}  \left [  \mathbf{P}_1 \mathbf{P}_2  \right ] \begin{bmatrix}
           \mathbf{g}_1 \\
           \mathbf{g}_2
         \end{bmatrix} =1.
\end{equation*}
After some manipulations the above equation leads to
\begin{equation}\label{eq:lem-prj-1}
\| \mathbf{g}_1 \|_2^2 + \| \mathbf{g}_2 \|_2^2 + 2 \mathbf{g}_1^T \mathbf{P}_1^T \mathbf{P}_2 \mathbf{g}_2 = 1.
\end{equation}
It suffices to have
\begin{equation}\label{eq:lem-prj-goal}
a+ 2\mathbf{g}_1^T \mathbf{P}_1^T \mathbf{P}_2 \mathbf{g}_2 + \mathbf{g}_2^T \mathbf{P}_2^T \mathbf{P}_1 \mathbf{P}_1^T \mathbf{P}_2 \mathbf{g}_2 + \mathbf{g}_1^T \mathbf{P}_1^T \mathbf{P}_2 \mathbf{P}_2^T \mathbf{P}_1 \mathbf{g}_1 \geq 0.
\end{equation}

\vspace{0.1in}

\noindent We write $\mathbf{P}_1^T \mathbf{P}_2 = \mathbf{Q} \Lambda \mathbf{V}^T$ as the singular value decomposition (svd) of $\mathbf{P}_1^T \mathbf{P}_2$. Note that orthogonal transformation will not change the $\ell_2$-norm, for ease of notation we simply write $\mathbf{g}_1 := \mathbf{Q}^T \mathbf{g}_1$ and $\mathbf{g}_2 := \mathbf{V}^T \mathbf{g}_2$. Let the diagonal of $\Lambda$ be $[ \lambda_1,...,\lambda_{r_2} ]$. We can expand~\eqref{eq:lem-prj-goal} and combine it with~\eqref{eq:lem-prj-1}, it suffices to have
\begin{align}\label{eq:lem-prj-2}
     & \sum_{i=1}^{r_2} (a+\lambda_i^2-\lambda_i)g_{1i}^2 + \sum_{i=1}^{r_2} (a+\lambda_i^2-\lambda_i)g_{2i}^2 + \sum_{i=1}^{r_2} \lambda_i (g_{1i}+g_{2i})^2 + 2 \sum_{i=1}^{r_2} a \lambda_i g_{1i} g_{2i} \geq 0.
\end{align}
Finally, we note that $a \geq \max_{1 \leq i \leq r_2}\lambda_i$, this means $a+\lambda_i^2-\lambda_i \geq \lambda_i^2$. Also from $a \leq \frac{1}{2}$ we know $a+\lambda_i^2-\lambda_i \geq a^2$, therefore~\eqref{eq:lem-prj-2} is true. Therefore the result is true for $n=2$.

\vspace{0.2in}

Now assume the result is true for $n$, we will prove the result is also true for $(n+1)$ by contradiction. Specifically, we can do Gram-Schmidt orthogonalization on columns of $[\mathbf{P}_1, ..., \mathbf{P}_n,\mathbf{P}_{n+1}]$ (it is fine to have $\mathbf{0}$ matrices, we just add empty columns at the end) and get $[\mathbf{\hat{P}}_1, ..., \mathbf{\hat{P}}_n,\mathbf{\hat{P}}_{n+1}]$. Write $\mathbf{g}=\sum_{i=1}^{n+1} \mathbf{g}_i$, here $\mathbf{g}_i$ is the projection of $\mathbf{g}$ onto $span(\mathbf{\hat{P}}_i)$. Then we know
\begin{align*}
    \sum_{i=1}^{n+1} \| \mathbf{P}_i^T \mathbf{g} \|_2^2 = \sum_{i=1}^{n+1} \| \mathbf{P}_i^T \mathbf{g}_i \|_2^2, \; \: \textit{and} \: \: \sum_{i=1}^{n+1} \| \mathbf{g}_i \|_2^2 = 1.
\end{align*}
By mathematical induction, the result follows naturally if we can show $\mathbf{g}_i = \mathbf{0}$ for some $i=1,..,n+1$. Now we assume $\| \mathbf{g}_i \|_2 >0$ for any $1 \leq i \leq n + 1$. If the result is not true for $(n+1)$, by induction we have
\begin{align*}
    &(1-\| \mathbf{g}_{n+1} \|_2^2)[1-(n-1)a] + \| \mathbf{P}_{n+1}^T \mathbf{g}_{n+1} \|_2^2 < 1-na \nonumber \\
    \Rightarrow & \| \mathbf{P}_{n+1}^T \mathbf{g}_{n+1} \|_2^2 < \| \mathbf{g}_{n+1} \|_2^2-a.
\end{align*}
Similarly we can get $\| \mathbf{P}_{i}^T \mathbf{g}_{i} \|_2^2 < \| \mathbf{g}_{i} \|_2^2-a$, for any $i=1,..,n,n+1$. This means
\begin{align*}
  [1-(n-1)a]\sum_{i=1}^n \| \mathbf{g}_{i} \|_2^2  \leq \sum_{i=1}^n \| \mathbf{P}_i^T \mathbf{g} \|_2^2 < \sum_{i=1}^n \| \mathbf{g}_{i} \|_2^2 - na \nonumber
     \Rightarrow  \sum_{i=1}^n \| \mathbf{g}_{i} \|_2^2 > \frac{n}{n-1},
\end{align*}
which is a contradiction! Hence the result is also true for $(n+1)$, and the lemma is proved.
$\hfill \blacksquare$

\newpage

\begin{lem}(See Lemma $8.4$ in~\cite{lerman2015robust} and Lemma $11$ in~\cite{rahmani2017coherence})\label{lem:lerman}
Let $\mathbf{g}_1,..,\mathbf{g}_n$ be i.i.d. random vectors from the unit sphere $\mathbb{S}^{r-1}$ and $r>1$. Then for all $t > 0$ we have
\begin{equation*}
    \inf_{\| \mathbf{v} \|_2 =1} \sum_{i=1}^n | \mathbf{v}^T \mathbf{g}_i | > \sqrt{\frac{2}{\pi}} \frac{n}{\sqrt{r}} - 2 \sqrt{n} - \sqrt{\frac{2n \log \frac{1}{t}}{r-1}},
\end{equation*}
and
\begin{equation*}
    \sup_{\| \mathbf{v} \|_2 =1} \sum_{i=1}^n | \mathbf{v}^T \mathbf{g}_i | < \frac{n}{\sqrt{r}} + 2\sqrt{n} + \sqrt{\frac{2n \log \frac{1}{t}}{r-1}},
\end{equation*}
both with probability at least $1-t$.
\end{lem}

\begin{lem}(See Lemma~1 in ~\cite{JMLR:v22:18-780})\label{lem:weiwei}
Let $\mathbf{b} \in \mathbb{R}^m$ sampled uniformly from $\mathbb{S}^{m-1}$, and $\lambda_k $ $(k=1,..,m)$ be constants such that $1 \geq \lambda_1 \geq \lambda_2 \geq ... \geq \lambda_m \geq 0$. For constant $g_1 \in (\lambda_m,\lambda_1)$, we write $r_i=(g_1^2-\lambda_i^2)_+$ and $s_i=(g_1^2-\lambda_i^2)_-$. Assume that $\sum_{i=1}^m r_i > \sum_{i=1}^m s_i$, then
\[
 \sum_{i=1}^m |\lambda_{i} b_{i}|^2 < g_1^2
\]
with probability at least $1-2 e^{-\epsilon^2}$, where
\[
\epsilon = \frac{ \sum_{i=1}^m(r_i-s_i)}{(\sqrt{\sum_{i=1}^m r_i^2}+\sqrt{\sum_{i=1}^m s_i^2})+\sqrt{(\sqrt{\sum_{i=1}^m r_i^2}+\sqrt{\sum_{i=1}^m s_i^2})^2+2 s_1 \sum_{i=1}^m(r_i-s_i)}}.
\]
\end{lem}

\vspace{0.1in}

With Lemma~1, Lemma~2, and Lemma~3 being true, we are now ready to prove Theorem~1.

\vspace{0.2in}

\noindent
\textbf{Proof of Theorem $1$}: Assume WLOG that $\mathbf{q}=\mathbf{d}_1^{(1)} \in \mathcal{S}_1$, we prove the iPursuit optimization problem~\eqref{eq:opt_mainn} has the same optimal solution with
\begin{align}\label{eq:op}
    & \min_{\mathbf{c} \perp \mathcal{S}_{-1}} \quad \left\| c^T \mathbf{D} \right\|_1,  \\
    & s.t. \quad \mathbf{c}^T \mathbf{q} =1. \nonumber
\end{align}
Let $c^*$, $c_2$ be the optimal solution of the iPursuit optimization problem and~\eqref{eq:op} associated with $\mathbf{q}$, respectively. Note that if iPursuit and~\eqref{eq:op} share the same optimal solution, all the points from other than $\mathcal{S}_1$ would be orthogonal to $c^{*}$. This means only points from $\mathcal{S}_1$ can have positive affinities with $\mathbf{d}_1^{(1)}$.

Following the proof in ~\cite{rahmani2017innovation}, it suffices to show
\begin{equation*}
    \left\| (\mathbf{c}_2-\mathbf{\delta})^T \mathbf{D} \right\|_1  -\left\| \mathbf{c}_2^T \mathbf{D} \right\|_1 >0
\end{equation*}
for any $\mathbf{\delta}^T \mathbf{q}$= 0. This can be reduced (see~\cite{rahmani2017innovation}) to be
\begin{align}\label{eq:bound}
    & \sum_{\mathbf{d}_i^{(-1)}}|\delta_1^T \mathbf{d}_i^{(-1)}|-\sum_{\mathbf{d}_i^{(1)}} |\delta_1^T \mathbf{d}_i^{(1)}| +\sum_{\mathbf{d}_i^{(1)}, i \in L_0} |\delta_I^T \mathbf{d}_i^{(1)}|- \sum_{\mathbf{d}_i^{(1)}, i \in L_0^c} sgn(\mathbf{c}_2^T \mathbf{d}_i^{(1)})\delta_I^T \mathbf{d}_i^{(1)} \geq 0,
\end{align}
where $\delta_I = (\mathbf{I}-\mathbf{U}_{-1}\mathbf{U}_{-1}^T) \delta$, $\delta_1 = \mathbf{U}_{-1}\mathbf{U}_{-1}^T \delta$, and $L_0 = \lbrace i \in [n_1]:\mathbf{c}_2^T \mathbf{d}_{i}^{(1)}=0 \rbrace$. The LHS of~\eqref{eq:bound} can be further lower bounded by
\begin{align}\label{eq:thm1-goal}
     \sum_{\mathbf{d}_i^{(-1)}}|\delta_1^T \mathbf{d}_i^{(-1)}|-\sum_{\mathbf{d}_i^{(1)}} |\mathbf{\delta}_1^T \mathbf{d}_i^{(1)}|-\sum_{\mathbf{d}_i^{(1)}} |\delta_{I2}^T \mathbf{d}_i^{(1)}|+ \sum_{\mathbf{d}_i^{(1)}, i \in L_0} |\delta_{I1}^T \mathbf{d}_i^{(1)}| - \sum_{\mathbf{d}_i^{(1)}, i \in L_0^c} sgn(\mathbf{c}_2^T \mathbf{d}_i^{(1)})\delta_{I1}^T \mathbf{d}_i^{(1)}.
\end{align}
Here we write $\delta_I$ into two parts $\delta_{I1}$ and $\delta_{I2}$ such that $\delta_{I1}$ is orthogonal to both $\mathbf{q}=\mathbf{d}_1^{(1)}$ and columns of $\mathbf{D}_{-1}$, and $\delta_{I2}=\delta_{I}-\delta_{I1}$.

Write $\delta_{I1} = (\mathbf{I}-\mathbf{Q} \mathbf{Q}^T) \delta_I$ and $\delta_{I2} = \mathbf{Q} \mathbf{Q}^T \delta_I$, where $\mathbf{Q}$ is the orthogonal basis of $span(\mathbf{D}_{-1}) \oplus span(\mathbf{q})$. From the definition of $\mathbf{c}_2$ we know
\begin{equation*}
    \sum_{\mathbf{d}_i^{(1)}, i \in L_0} |\delta_{I1}^T \mathbf{d}_i^{(1)}| - \sum_{\mathbf{d}_i^{(1)}, i \in L_0^c} sgn(\mathbf{c}_2^T \mathbf{d}_i^{(1)})\delta_{I1}^T \mathbf{d}_i^{(1)}\geq 0.
\end{equation*}
Note that inequality~\eqref{eq:bound} is true follows trivially with $\delta_1 = 0$.  Now we are going to bound the terms in~\eqref{eq:bound} with $\delta_1 \neq 0$. Specifically, we will prove the sum of the first three terms at the LHS of~\eqref{eq:thm1-goal} is non-negative.

Write $\delta_{1k}=\mathbf{U}_k^T\delta_1$ we have
\begin{align*}
    \sum_{\mathbf{d}_i^{(-1)}} | \delta_1^T \mathbf{d}_i^{(-1)}| &= \sum_{k=2}^K \sum_{\mathbf{b}_i^{(k)}} | \delta_{1k}^T \mathbf{b}_i^{(k)} | \nonumber \\
    & \geq \sum_{k=2}^K \| \mathbf{U}_k^T \delta_1 \|_2 \cdot \inf_{ \|\hat{\delta}_{1k}\|_2=1} \sum_{\mathbf{b}_i^{(k)}} | \hat{\delta}_{1k}^T \mathbf{b}_i^{(k)} | \nonumber \\
   &\geq \left [ \min_{k} \inf_{ \|\hat{\delta}_{1k}\|_2=1} \sum_{\mathbf{b}_i^{(k)}} | \hat{\delta}_{1k}^T \mathbf{b}_i^{(k)} | \right ] \sum_{k=2}^K \| \mathbf{U}_k^T \delta_1 \|_2 \nonumber \\
   & = h_1 \| \delta_1 \|_2 \sum_{k=2}^K \| \mathbf{U}_k^T \hat{\delta}_1 \|_2,
\end{align*}
where $\hat{\delta}_1$ is normalized $\delta_1$. We can write $\hat{\delta}_1 =  \delta_{1U} + \delta_{1C} $ such that $\delta_{1U}$ is the projection of $\hat{\delta}_1$ onto $span(\mathbf{U})$, then we have $\| \delta_{1U} \|_2^2 + \| \delta_{1C} \|_2^2=1$ and
\begin{align*}
    \| \mathbf{U}_k^T \hat{\delta}_1 \|_2^2 = \| \delta_{1U} \|_2^2 + \| \hat{\mathbf{U}}_k^T \delta_{1C}\|_2^2.
\end{align*}
Therefore
\begin{align}\label{eq:thm1-term1}
    \sum_{k=2}^K \| \mathbf{U}_k^T \hat{\delta}_1 \|_2^2 &= (K-1) \| \delta_{1U} \|_2^2 +  \sum_{k=2}^K \| \mathbf{\hat{U}}_k^T \delta_{1C} \|_2^2 \nonumber \\
    & \geq K-1 - \left [ K-2+(K-2)t_2 \right ]\|\delta_{1C} \|_2^2,
\end{align}
the inequality above comes from Lemma~\ref{lem:prj}. Put everything together we have
\begin{equation*}
    \sum_{\mathbf{d}_i^{(-1)}} | \delta_1^T \mathbf{d}_i^{(-1)}| \geq h_1 \|\delta_1 \|_2 \sqrt{K-1 - \left [ K-2+(K-2)t_2 \right ]\|\delta_{1C} \|_2^2}.
\end{equation*}

Now consider the term $\sum_{\mathbf{d}_i^{(1)}}|\delta_1^T \mathbf{d}_i^{(1)}|$. Similarly as before we have
\begin{align*}
    \sum_{\mathbf{d}_i^{(1)}}| \delta_1^T \mathbf{d}_i^{(1)}| &= \| \mathbf{U}_1^T \delta_1 \|_2 \cdot \sum_{\mathbf{b}_i^{(1)}}  \frac{| \delta_1^T \mathbf{U}_1 \mathbf{b}_i^{(1)}|}{\| \mathbf{U}_1^T \delta_1 \|_2 } \leq \| \mathbf{U}_1^T \delta_1 \|_2 \cdot \sup_{\| \hat{\delta}_{11} \|_2 =1} \sum_{\mathbf{b}_i^{(1)}} | \hat{\delta}_{11}^T \mathbf{b}_i^{(1)} |,
\end{align*}
and
\begin{align*}
    \| \mathbf{U}_1^T \hat{\delta}_{1} \|_2^2 &= \| \delta_{1U} \|_2^2 + \| \mathbf{\hat{U}}_1^T \delta_{1C} \|_2^2 \leq 1 - \| \delta_{1C} \|_2^2 + t_1^2 \| \delta_{1C} \|_2^2.
\end{align*}
Therefore we have
\begin{align}\label{eq:thm1-term2}
    \sum_{\mathbf{d}_i^{(1)}}| \delta_1^T \mathbf{d}_i^{(1)}| \leq  \sqrt{1+(t_1^2-1)\|\delta_{1C} \|_2^2} \cdot \| \delta_1\|_2 \cdot h_2.
\end{align}

Finally we consider $\| \mathbf{U}_{1}^T \delta_{I2} \|_2$. Write the orthogonal basis $\mathbf{Q} = \left [ \mathbf{U}_{-1}, \mathbf{\hat{q}} \right ] $, here $\mathbf{\hat{q}} = \frac{(\mathbf{I}-\mathbf{U}_{-1}\mathbf{U}_{-1}^T)\mathbf{q}}{\| (\mathbf{I}-\mathbf{U}_{-1}\mathbf{U}_{-1}^T)\mathbf{q} \|_2}$. The case where $\hat{\mathbf{q}} = \mathbf{0}$ is trivial, hence we assume $\hat{\mathbf{q}} \neq \mathbf{0}$. Note that
\begin{align}
    \| \mathbf{U}_1^T \delta_{I2} \|_2 & \leq \| \delta_{I2} \|_2 \nonumber \\
    & = \| (\mathbf{Q} \mathbf{Q}^T - \mathbf{U}_{-1} \mathbf{U}_{-1}^T)\delta \|_2 \nonumber \\
    & = \left \| \mathbf{\hat{q}} \frac{\mathbf{q}^T(\delta-\mathbf{U}_{-1}\mathbf{U}_{-1}^T \delta)}{\| (\mathbf{I}-\mathbf{U}_{-1}\mathbf{U}_{-1}^T)\mathbf{q} \|_2} \right \|_2 \nonumber \\
    & = \left \| \frac{\mathbf{\hat{q}} \mathbf{q}^T \delta_1}{\| (\mathbf{I}-\mathbf{U}_{-1}\mathbf{U}_{-1}^T)\mathbf{q} \|_2} \right\|_2 \nonumber \\
    & \leq \sqrt{\frac{t_3^2}{1-t_3^2}}\cdot \| \delta_1 \|_2. \nonumber
\end{align}
Hence we have
\begin{equation}\label{eq:thm1-term3}
    \sum_{\mathbf{d}_i^{(1)}} |\delta_{I2}^T \mathbf{d}_i^{(1)}| \leq \sqrt{\frac{t_3^2}{1-t_3^2}}\cdot \| \delta_1 \|_2 h_2,
\end{equation}
where the last inequality comes from our assumption.

Combing~\eqref{eq:thm1-term1},~\eqref{eq:thm1-term2} and~\eqref{eq:thm1-term3}, it suffices to show
\begin{align*}
    h_1 \sqrt{ K-1 - \left [ K-2+(K-2)t_2 \right ]\|\delta_{1C} \|_2^2 } \geq h_2 \left (\sqrt{\frac{t_3^2}{1-t_3^2}}+ \sqrt{\left [ 1+(t_1^2-1)\|\delta_{1C} \|_2^2 \right ] } \right).
\end{align*}
We only need the above inequality to be true for both $\|\delta_{1C} \|_2=0$ and $\|\delta_{1C} \|_2=1$, which means it suffices to have the following two inequalities
\begin{align*}
     h_1 \sqrt{ 1-(K-2)t_2} \geq h_2 \left(\sqrt{\frac{t_3^2}{1-t_3^2}}+ t_1 \right),\quad h_1 \sqrt{K-1} \geq h_2 \left(\sqrt{\frac{t_3^2}{1-t_3^2}}+ 1 \right).
\end{align*}
Both of them are true based on assumption.
$\hfill \blacksquare$

\vspace{0.2in}

\noindent
\textbf{Proof of Theorem $2$}:
From Lemma~\ref{lem:lerman} we know that for any $\mathbf{v} \in \mathbb{R}^m$ the following inequalities hold
\begin{equation*}
    \inf_{\| \mathbf{v} \|_2 =1 } \sum_{i=1}^{n} | \mathbf{v}^T \mathbf{b}_i^{(k)} | > \sqrt{\frac{2}{\pi}} \frac{n}{\sqrt{m}} - 2 \sqrt{n} - \sqrt{\frac{2n \log n}{m-1}},
\end{equation*}
and
\begin{equation*}
    \sup_{\| \mathbf{v} \|_2 =1} \sum_{i=1}^{n} | \mathbf{v}^T \mathbf{b}_i^{(k)} | < \frac{n}{\sqrt{m}} + 2\sqrt{n} + \sqrt{\frac{2n \log n}{m-1}},
\end{equation*}
both with probability at least $1-\frac{1}{n}$.

Again we assume $\mathbf{q} \in \mathcal{S}_1$ and write $\mathbf{q}= \left [ \mathbf{U},\mathbf{\hat{U}}_1 \right ] \mathbf{\hat{q}}$, here $\mathbf{\hat{q}} \in \mathbb{R}^m$ is a sample from the uniform distribution on $\mathbb{S}^{m-1}$. For an arbitrary point $\delta \in \mathcal{S}_{-1}$ with unit norm, we can write $\delta = \left [ \mathbf{U}, \mathbf{\Tilde{U}}_2,...,\mathbf{\Tilde{U}}_K \right ] \hat{\delta}$, here $\mathbf{\Tilde{U}}_k$ is the orthogonal basis for the innovation directions in $\mathcal{S}_k$ with respect to all other subspaces minus $span(\mathbf{\hat{U}}_1)$ (we can add zero columns to guarantee $\mathbf{\tilde{{U}}}_k$ has same dimension as $\mathbf{\hat{U}}_k$). We then have the following relations
\begin{align}
    |\mathbf{q}^T \delta |^2 &= \left| \hat{\mathbf{q}}^T   \cdot
        \begin{bmatrix}
           \mathbf{U}^T \\
           \mathbf{\hat{U}}_1^T
        \end{bmatrix} \left [ \mathbf{U}, \mathbf{\Tilde{U}}_2,...,\mathbf{\Tilde{U}}_K \right ] \hat{\delta} \right|^2 \nonumber \\
        & = \left| \mathbf{\hat{q}}^T \begin{bmatrix}
           \mathbf{I},\quad \mathbf{0},......,\quad \mathbf{0} \\
           \mathbf{0},\mathbf{\hat{U}}_1^T \mathbf{\Tilde{U}}_2,..,\mathbf{\hat{U}}_1^T \mathbf{\Tilde{U}}_K
        \end{bmatrix} \hat{\delta} \right|^2\nonumber \\
        & \leq  \sum_{i=1}^s \hat{q}_i^2 + (K-1)t_2^2 \sum_{i=s+1}^m \hat{q}_i^2. \nonumber
\end{align}
Now we write $\hat{q}_i^2 = \frac{z_i^2}{\sum_{i=1}^m z_i^2}$, where $z_i$'s are i.i.d. standard Gaussian random variables. For positive constant $t_3$ we have
\begin{align*}
    & \mathbb{P}\left [ \sum_{i=1}^s \hat{q}_i^2 + (K-1)t_2^2 \sum_{i=s+1}^{m}\hat{q}_i^2 > t_3^2 \right ] \\
    = &  \mathbb{P} \left [  (1-t_3^2)\sum_{i=1}^s z_i^2 > \left[ t_3^2-(K-1)t_2^2 \right]\sum_{i=s+1}^m z_i^2 \right] \\
    = &  \mathbb{P} \left [ \frac{\sum_{i=1}^s z_i^2/s }{\sum_{i=s+1}^m z_i^2/(m-s)} > \frac{\left[ t_3^2-(K-1)t_2^2 \right]}{1-t_3^2} \cdot \frac{m-s}{s} \right]
    \leq  2 e^{-\epsilon^2},
\end{align*}
where from Lemma $1$ in~\cite{JMLR:v22:18-780} and our assumption in~\eqref{eq:new_int2} we know
\begin{equation*}
   \epsilon=\frac{1}{2}\left[ -(\sqrt{s}+\frac{\zeta s}{\sqrt{m-s}})\sqrt{(\sqrt{s}+\frac{\zeta s}{\sqrt{m-s}})^2+2s(\zeta-1)}     \right],
\end{equation*}
and $\zeta = \frac{(m-s)\left[ t_3^2-(K-1)t_2^2 \right]}{s(1-t_3^2)} >1 $. We conclude the proof by using union bound inequality.
$\hfill \blacksquare$
\vspace{0.2in}

\noindent
Define $\theta_k^{(1)}$ as the smallest principal angle between $\mathit{span}(\mathbf{\hat{U}}_k)$ and $\mathit{span}(\mathbf{\hat{U}}_{-k})$. Then from~\cite{johnstone2008multivariate} we know $\cos^2{\theta_k^{(1)}}$ is the largest root of a multivariate beta distribution. We introduce the following lemma that can be used to approximate the true value of $\cos^2{\theta_k^{(1)}}$.
\begin{lem}\label{lem:johnstone}(\cite{johnstone2008multivariate})
Assume $(K-1)(m-s)+s$ is even and
\begin{equation*}
    \lim_{m \rightarrow \infty} \frac{m-s}{M} >0, \quad \lim_{m \rightarrow \infty} \frac{(K-1)(m-s)+s}{M-m+s} < 1.
\end{equation*}
Then $\cos^2{\theta_k^{(1)}} = \mu + \sigma F_1 + O(m^{-4/3})$, where
\begin{align*}
    & \mu = \sin^2{\frac{\phi+\gamma}{2}}, \quad \sigma^3 = \frac{\sin^4(\phi + \gamma)}{4(M-1)^2\sin{\phi}\sin{\gamma}}, \\
    & \frac{\gamma}{2} = \sqrt{\frac{(K-1)(m-s)+s}{M}}, \quad  \frac{\phi}{2} = \sqrt{\frac{m-s}{M}},
\end{align*}
here $F_1$ is a random variable that follows the limiting distribution of the largest eigenvalue of a $\left [ (K-1)(m-s)+s \right ]\times \left [ (K-1)(m-s)+s \right ]$ Gaussian symmetric matrix.
\end{lem}
Based on Lemma~4, The proof of Theorem $3$ is then a combination of that of Theorems $1$ and $2$ and is omitted here.

\vspace{0.3in}

\noindent
\textbf{Proof of Theorem $4$}:
 We know $\mathbf{X}_{[:,1]}$ is the solution of the following optimization problem
\begin{equation}\label{eq:eigen}
    \max_{\| \mathbf{v} \|_2=1} \mathbf{v}^T \mathbf{D} \mathbf{D}^T \mathbf{v}.
\end{equation}
Now write $\mathbf{v}^* = \mathbf{X}_{[:,1]}$, by the definition of $\mathbf{v}^*$ we know
\begin{align}\label{eq:inter_to_inno}
    &\|\lambda_1 \mathbf{v}_u^T \mathbf{D}+ \lambda_2 \mathbf{v}_c^T \mathbf{D}\|_2 \geq \| \mathbf{v}_u^T \mathbf{D} \|_2 \nonumber \\
    \Rightarrow & \|\mathbf{v}_c^T \mathbf{D}\|_2^2 + \frac{2\lambda_1}{\lambda_2} \|\mathbf{v}_u^T D\|_2 \| \mathbf{v}_c^T D \|_2 \geq \| \mathbf{v}_u^T D \|_2 ^2.
\end{align}
Now we can roughly estimate $\frac{\| \mathbf{v}_u^T D \|_2 ^2}{\|\mathbf{v}_c^T D\|_2^2}$. Write $\mathbf{v}_u = \mathbf{U} \mathbf{\hat{v}}_u $, $\mathbf{v}_c = \mathbf{U}_c \mathbf{\hat{v}}_c $ and $\mathbf{d}_i^{(k)} = \mathbf{U} \mathbf{b}_{iu}^{(k)} + \mathbf{\hat{U}}_k \mathbf{b}_{ic}^{(k)}$, here $\begin{bmatrix}
           \mathbf{b}_{iu}^{(k)} \\
           \mathbf{b}_{ic}^{(k)}
         \end{bmatrix}$ has unit norm. Then note
\begin{align*}
    &\| \mathbf{v}_u^T D \|_2 ^2 = \sum_{k=1}^K \sum_{i=1}^{n} (\mathbf{v}_u^T \mathbf{d}_i^{k})^2 = \sum_{k=1}^K \sum_{i=1}^{n} (\mathbf{\hat{v}}_u^T \mathbf{b}_{iu}^{(k)})^2,  \\
    &\| \mathbf{v}_c^T D \|_2 ^2 = \sum_{k=1}^K \sum_{i=1}^{n} (\mathbf{v}_c^T \mathbf{d}_i^{k})^2 = \sum_{k=1}^K \sum_{i=1}^{n} (\mathbf{\hat{v}}_c^T \mathbf{\hat{U}}_k \mathbf{b}_{ic}^{(k)})^2.
\end{align*}

We start from $\sum_{k=1}^K \sum_{i=1}^{n} (\mathbf{\hat{v}}_u^T \mathbf{b}_{iu}^{(k)})^2$. By Cauchy-Schwartz inequality we have:
\begin{align*}
    Kn \cdot \sum_{k=1}^K \sum_{i=1}^{n} (\mathbf{\hat{v}}_u^T \mathbf{b}_{iu}^{(k)})^2 &\geq \left [ \sum_{k=1}^K \sum_{i=1}^{n} |\mathbf{\hat{v}}_u^T \mathbf{b}_{iu}^{(k)}| \right ]^2= \left [ \sum_{k=1}^K \sum_{i=1}^{n} \left |
    \begin{bmatrix}
           \mathbf{\hat{v}}_u^T,\mathbf{0} \\
         \end{bmatrix} \begin{bmatrix}
           \mathbf{b}_{iu}^{(k)} \\
           \mathbf{b}_{ic}^{(k)}
         \end{bmatrix} \right | \right ]^2.
\end{align*}
From Lemma~\ref{lem:lerman} we know
\begin{equation*}
    p\left[ \sum_{k=1}^K \sum_{i=1}^{n} (\mathbf{\hat{v}}_u^T \mathbf{b}_{iu}^{(k)})^2 \geq  \left[ \sqrt{ \frac{2Kn}{\pi m}} - 2 - \sqrt{\frac{2\log n}{m-1}} \right ]^2 \right] \geq 1-\frac{1}{n}.
\end{equation*}
It is straightforward to see that for large $s$ (i.e., $s \approx m$), the above inequality is tighter since $\| \mathbf{b}_{iu}^{(k)} \|_2 \approx 1$, the cost of ``losing'' $\mathbf{b}_{ic}^{(k)}$ is small.

\newpage

As for $\sum_{k=1}^K \sum_{i=1}^{n} (\mathbf{v}_c^T \mathbf{\hat{U}}_k \mathbf{b}_{ic}^{(k)})^2$, similarly from Cauchy-Schwartz inequality and Lemma~\ref{lem:prj} we have
\begin{align*}
    \sum_{k=1}^K \sum_{i=1}^{n} (\mathbf{v}_c^T \mathbf{\hat{U}}_k \mathbf{b}_{ic}^{(k)})^2  \leq \sum_{k=1}^K \| \mathbf{\hat{U}}_k^T \mathbf{v}_c \|_2^2 \max_{k} \sum_{i=1}^{n}  \| \mathbf{b}_{ic}^{(k)} \|_2^2
     \leq \left [ 1+(K-1)t \right ] \max_{k} \sum_{i=1}^{n}  \| \mathbf{b}_{ic}^{(k)} \|_2^2.
\end{align*}
For some fixed $k$ and $i$, we can write $ \| \mathbf{b}_{ic}^{(k)} \|_2^2 = \frac{\sum_{i=1}^{m-s} z_i^2}{\sum_{i=1}^m z_i^2}$, here $z_i \sim \mathcal{N}(0,1)$.

\vspace{0.1in}

Then from Lemma~\ref{lem:weiwei} we know for positive constant $\kappa \in (\frac{m-s}{m},1)$:
\begin{align*}
    P \left[ \| \mathbf{b}_{ic}^{(k)} \|_2^2 \geq \kappa \right] &= P \left[ \frac{\sum_{i=1}^{m-s} z_i^2}{\sum_{i=1}^m z_i^2} \geq \kappa\right]  \\
    & \leq 2e^{-\epsilon^2},
\end{align*}
where
\[
\epsilon = \frac{(m-s)\left [ \frac{\kappa s }{(1-\kappa)(m-s)}-1 \right ]}{(\sqrt{m-s}+\frac{\kappa\sqrt{s}}{1-\kappa})+\sqrt{(\sqrt{m-s}+\frac{\kappa\sqrt{s}}{1-\kappa})^2+2(m-s)\left[  \frac{\kappa s }{(1-\kappa)(m-s)}-1 \right]}}.
\]
For large $s$ and if $m$ and $s$ are in linear order with each other, $\epsilon$ is roughly in the order of $\sqrt{m}$.

\vspace{0.1in}

Therefore from union bound inequality we have
\begin{align*}
    \frac{\sum_{k=1}^K \sum_{i=1}^{n} (\mathbf{\hat{v}}_u^T \mathbf{b}_{iu}^{(k)})^2}{\sum_{k=1}^K \sum_{i=1}^{n} (\mathbf{v}_c^T \mathbf{\hat{U}}_k \mathbf{b}_{ic}^{(k)})^2} \geq \frac{m\left[ \sqrt{ \frac{2Kn}{\pi m}} - 2 - \sqrt{\frac{2\log n}{m-1}} \right ]^2 }{\left[1+(K-1)t \right]n(m-s)\kappa},
\end{align*}
with probability at least $1-\frac{1}{n}-2Kne^{-\epsilon^2}$.
$\hfill \blacksquare$

\end{document}